\documentclass[letterpaper, 10 pt, conference]{ieeeconf}  

\IEEEoverridecommandlockouts                              

\usepackage{cite}
\usepackage{graphics} 
\usepackage{epsfig} 
\usepackage{mathptmx} 
\usepackage{times} 
\usepackage{amsmath} 
\usepackage{amssymb}  
\usepackage{color}
\usepackage{blkarray}
\usepackage{mathtools}
\usepackage{breqn}
\usepackage{algorithmicx}
\usepackage{algorithm}
\usepackage[noend]{algpseudocode}
\usepackage{stfloats}
\usepackage{subfig}
\usepackage{svg}
\usepackage{url}

\makeatletter
\def\BState{\State\hskip-\ALG@thistlm}
\makeatother

\DeclareMathOperator*{\argmax}{arg\,max}

\title{\LARGE \bf
Autonomous Skill-centric Testing using Deep Learning
}

\author{Simon Hangl, Sebastian Stabinger and Justus Piater$^{1}$
\thanks{$^{1}$ All authors are with the Department of Computer Science,
		Universit\"at Innsbruck, 6020 Innsbruck, Austria
		{\tt\small firstname.lastname@uibk.ac.at}}%
}

\begin{document}

\maketitle
\thispagestyle{empty}
\pagestyle{empty}

\begin{abstract}
Software testing is an important tool to ensure software quality.
This is a hard task in robotics due to dynamic environments
and the expensive development and time-consuming execution of test cases.
Most testing approaches use model-based and$\,$/$\,$or simulation-based
testing to overcome these problems.
We propose model-free skill-centric testing in which a robot autonomously
executes skills in the real world and compares it to previous experiences.
The skills are selected by maximising the expected information gain on the distribution of
erroneous software functions.
We use deep learning to model the sensor data observed during previous
successful skill executions and to detect irregularities.
Sensor data is connected to function call profiles such that
certain misbehaviour can be related to specific functions.
We evaluate our approach in simulation and in experiments
with a KUKA LWR 4+ robot by purposefully introducing bugs to the software.
We demonstrate that these bugs can be detected with high accuracy
and without the need for the implementation of specific tests or
task-specific models.
\end{abstract}

\section{Introduction}
\label{sec:introduction}
In recent years robot programming increasingly matured and trained skills became more complex.
Programming paradigms shifted from pure hard-coding of skills to (semi-) autonomous
skill acquisition techniques.
Human supervisors are asked for advice only when needed,
e.g. in \emph{programming by demonstration}
and/or reinforcement learning \cite{Argall2009469, cypher1993watch, Hangl-2015-ICAR}.
These new paradigms typically produce a large corpus of mostly unused data.
Further, skills involve synchronisation of a high number of components (e.g. computer vision,
navigation, path planners, force/torque sensors, artificial skins) and are
implemented by large teams of experts in their respective fields.
Software components change rapidly which makes
it very hard for one single programmer to debug certain components in case of failure.
This problem becomes even more severe if the developer is not an expert in the field
to which a certain component belongs.
Currently, software testing is done in simulation with a high initial cost of
implementing test cases.
This has some obvious advantages such as the high repeatability
and higher safety.
However, this testing paradigm also has some serious disadvantages
such as the limited conformity with the world or
the demand for precise models of the physical environment.

We propose a novel testing scheme that does not require
the test engineer to be an expert.
Skills are executed autonomously and
all collected data, i.e.\ sensor data and function call profiles,
is compared to previous experiences.
From this stack of experiences two models can be trained:
a \emph{measurement observation model} (MOM) and a \emph{functional profiling fingerprint} (FPF).
The MOM models sensor measurements observed during successful skill executions.
It is trained with deep learning and, as opposed to other testing approaches in robotics, does
not impose a strong prior on the nature of tasks or functions to be tested.
Therefore our approach is model-free in a sense that no component-specific models are predefined.
The FPF provides a typical fingerprint of function calls for a certain skill.
The idea is to identify the point in time, in which a skill execution failed, by using the MOM
and to relate this to anomalies in the FPF.
The skills are selected in order to maximise the expected information
gain about which functions might cause problems.
The suggestions about possible erroneous functions
can be forwarded to the respective developers.

We do not distinguish between skills and tests, which eliminates the need
for developing specific test cases.
Currently, experts spend time to design test cases
with well-defined pre- and post conditions.
They typically also do not have a system overview
which causes the test cases to be too loose or too strict.
In our paradigm, the test cases are just as strict as they
need to be as they are grounded on the skills.

The concept is illustrated by a task in which 
an object is to be pushed along a given path.
Many components such as robot control software, an object tracker
(image acquisition, segmentation, localisation),
a path planner and pushing-specific functions are involved.
If the robot loses contact to the object it will
be detected by comparing the end-effector force measurements
to previous experiences.
Our approach will first suggest all running functions from
segmentation to pushing-specific functions.
If the scene segmentation contains a bug, all functions not related
to segmentation can be eliminated as a cause
by running skills that use a different segmentation
algorithm or no object tracking at all.
\section{Related Work}
Bihlmaier et al.\ propose robot unit testing \emph{RUT} \cite{Bihlmaier2014},
closely relating to the idea of classical unit tests.
They rely on simulators that are sufficiently accurate
and argue that this indeed is the case for most tasks.
Laval et al.\ use pre-defined tests for robot hardware
in an industrial setting, in which robots are manufactured
in high numbers \cite{laval01}.
In such a scenario, testing can only be performed in an autonomous manner.
They provide a multi-layered testing approach with
guidelines on how to test hardware with hand-crafted test cases
and how to share them between quality assurance and maintenance staff.
A similar idea is proposed by Lim et al.\ by using a hierarchical
testing framework composed of unit testing, integration testing
and system testing \cite{lim2010automated}.
As opposed to this work, these approaches are guidelines
for the developers rather than testing frameworks for autonomous robots.
Regression testing is applied to robotics by Biggs \cite{biggs01}.
He argues that even testing of low-level control stacks should
be performed in simulation in order to prevent dangerous situations.
Previous communication of network-based control middle-ware is stored and
fed back for testing again.
Even though this approach is similar in nature to our framework,
it is restricted to and designed for low-level components.
Zaman and Steinbauer et al.\ describe a diagnosis framework and augment it with
the additional ability of autonomous self-repair
\cite{zaman2013, steinbauer2005detecting}.
The framework is embedded into the ROS diagnostics stack
in which single components can publish observations
and diagnostic messages.
It requires pre-defined abstract diagnosis
models, whereas our method learns the model from experience.
Petters et al.\ provide a set of tools to ensure
the proper working of the control software used by teams
of autonomous mobile robots \cite{Petters2008}.
This involves a high level of manual test design.

Simulation-based testing \cite{son_kuc, park2012sitaf}
is one of the most widespread testing techniques.
Son et al.\ propose a simulation-based framework and guidelines
for unit, state, and interface testing, which involves the
generation of test cases and the execution in a simulator \cite{son_kuc}.
Park and Kang propose the SITAF architecture  \cite{park2012sitaf}
for testing software components in which
tests are specified as abstractly as possible.
Tests are generated automatically and are run on a simulator.
Related work is also done in the area of fault detection
\cite{frank1997survey, hashimoto2003multi, verma2004real,
1501236, 1570640, 4375131, 4399594, kawabata2003system,
ROB:ROB1029}.
Many of these systems follow an observer-based approach in which
separate observers monitor components.
Reasoning over the observed data in combination with a
\emph{pre-defined} model allows failures to be identified in dynamic systems.
An extensive discussion of fault detection systems is outside the
scope of this work, as these methods are mainly concerned with model-based
approaches applied to specific scenarios or components.

Work that also relies on automatic data storage from previous successful
experiences was proposed by Niemueller et al. \cite{6385940}.
Data is queried automatically during the execution of skills and stored
to a database.
It is directly taken from listening to ROS topics and stored
to the NoSQL database \emph{MongoDB}.
They demonstrate the applicability of automatic data storage
to fault analysis by hand-crafting a \emph{Data-Information-Knowledge-Wisdom}
hierarchy \cite{rowley2007wisdom}, which represents different levels
of abstraction.
Developers can then manually work through the hierarchy and identify potential
errors by comparing current sensor data on several levels to previous experiences.
This demonstrates that such an approach makes sense in principle,
however, in our method the identification of problems is done completely
autonomously.
\section{Skill-centric Autonomous Testing}
\label{sec:method}
A skill is a pair of a state-changing behaviour
and a predicate that determines success.
A \emph{behaviour} is a function
\begin{equation}
	b : S \mapsto S
\label{eq:behaviour}
\end{equation}
that maps an environment state $\mathbf{s} \in S$
to another state $\mathbf{s'} \in S$.
A \emph{skill} $a = (b, \beta)$ consists of a behaviour $b \in B$
and a predicate
\begin{equation}
	\beta \left( b \left( \mathbf{s} \right) \right) = \text{true}
\end{equation}
with $\mathbf{s} \in D$.
The predicate $\beta$ provides a notion of success for the behaviour $b$.
The set $D \subseteq S$ is called the \emph{domain of applicability} of the skill $a$.
We assume the robot holds a set of well-trained skills $A$,
i.e. the set $D_{a}$ is large with $|D_{a}| \gg 1$.

For each skill $a \in A$, the robot holds a \emph{database} $\Delta_a$ of pairs
\begin{equation}
	\Delta_a = \{ \left( \mathbf{M}_a(\mathbf{s}), \, \mathbf{F}_a(\mathbf{s}) \right) \}
\end{equation}
These pairs are \emph{positive experiences} of successful skill executions,
i.e. $\beta(\mathbf{s}) = \text{true}$.
The matrix $\mathbf{M}_a(\mathbf{s})$ contains the sensor data
(e.g. $\mathbf{m}_1 = $ force/torque sensors,
$\mathbf{m}_2 = $ position data, $\mathbf{m}_3 = $ images, \dots)
measured during execution of $a$ by
\begin{equation}
\mathbf{M}_a(\mathbf{s}) =
\begin{blockarray}{cccc}
	\begin{block}{(ccc)c}
		\mathbf{m}_1(0) & \dots & \mathbf{m}_1(T) & \text{sensor } 1 \\
		\mathbf{m}_2(0) & \dots & \mathbf{m}_2(T) & \text{sensor } 2 \\
		\vdots & \ddots & \vdots & \downarrow \\
		\mathbf{m}_M(0) & \dots & \mathbf{m}_M(T) & \text{sensor } M  \\
	\end{block}
	t = 1 & \xrightarrow{\Delta t} & t = T & \\
\end{blockarray}
\end{equation}
with $dim(\mathbf{M}_a(\mathbf{s})) = \left( \sum_i{dim(\mathbf{m}_i)}, T \right)$
and the execution time $T$.
Analogously, the \emph{profiling matrix} $\mathbf{F}_a(\mathbf{s})$ is given by
\begin{equation}
\mathbf{F}_a(\mathbf{s}) =
\begin{blockarray}{cccc}
	\begin{block}{(ccc)c}
		fc_1(0) & \dots & fc_1(T) & \text{function } 1 \\
		fc_2(0) & \dots & fc_2(T) & \text{function } 2 \\
		\vdots & \ddots & \vdots & \downarrow \\
		fc_F(0) & \dots & fc_F(T) & \text{function } F  \\
	\end{block}
	t = 1 & \xrightarrow{\Delta t} & t = T & \\
\end{blockarray}
\label{equ:fingerprintmatrix}
\end{equation}
where $fc_i(t)$ denotes the number of active executions of function $i$ at time
$[t, \, t + 1]$.
Two distributions can be estimated from $\Delta_a$:
the \emph{measurement observation model} (MOM)
\begin{equation}
p_a \left( succ \, | \, \mathbf{M}, \, t \right)
\end{equation}
and the \emph{functional profiling fingerprint} (FPF)
\begin{equation}
p_a^f \left( fc \, | \, \mathbf{M}, \, t, \, succ = \text{true} \right)
\end{equation}
The MOM reflects the probability of a successful execution of skill $a$
given the sensor data $\mathbf{M}$ at time $t$.
The FPF denotes the probability of how many instances of a function $f$
were active in the time period $[t, \, t + 1]$.
The conditioning on $\mathbf{M}$ is required in order to include
closed-loop controllers that can chose certain actions dependent
on the current measurement.

The goal is to execute skills from the robot's skill repertoire in order to
find a \emph{blaming distribution}
\begin{equation}
	p_{\text{blame}} \left( f = f_i \, | \, o_{1:T}, \, a_{1:T} \right)
\end{equation}
given a sequence of executed skills $a_{1:T}$ and corresponding
observations $o_{1:T}$.
Each observation $o = \left( \mathbf{M}, \, \mathbf{F} \right)$ contains sensor data and a functional
profile. The skills are selected by maximising the expected information gain
given the current belief $p_{\text{blame}}$ (section \ref{sec:informationgain}).
After each execution of a skill, the belief is updated by Bayesian inference
(section \ref{sec:bayesian}).
The complete algorithm is summarised in Algorithm \ref{alg:overall}.
\begin{algorithm}
\caption{Algorithm for autonomous skill-based testing}
\label{alg:overall}
\begin{algorithmic}[1]
\State Uniformly initialise
$p_0 = p_{\text{blame}} \left( f \, | \, \emptyset, \, \emptyset \right)$
\State Compute information gain and next skill $(a_0, I_0) \leftarrow$ \Call{Optimise IG}{$p_0$}
\State Initialise $t \leftarrow 0$
\While{$I_t$ not converged}
	\State $(a_{t + 1}, I_{t + 1}) \leftarrow$ \Call{Optimise IG}{$p_t$}
	\State Execute $a_{t + 1}$ and observe $o_{t + 1} = (\mathbf{M}_{a_{t+1}}, \mathbf{F}_{a_{t+1}})$
	\If{$succ_{t + 1} \neq \text{true}$}
		\State Use MOM to estimate $t_{\text{fail}}$
	\Else
		\State $t_{\text{fail}} \leftarrow$ not required
	\EndIf
	\State $p_{t + 1} \leftarrow$ \Call{Bayes}{$p_t$, $o_{t + 1}$, $a_{t + 1}$, $succ_{t + 1}$, $t_{\text{fail}}$}
	\State $t \leftarrow t + 1$
\EndWhile
\Function{Optimise IG}{$p_{\text{blame}}\left( f \, | \, o_{1:t}, \, a_{1:t} \right)$}
	\State Compute $H \left[ p_{\text{blame}} \right]$
	\For{\textbf{each} skill $a \in A$}
		\State Initialise set of sampled entropies $H_{\text{sam}} \leftarrow \emptyset$
		\For{\textbf{each} observation $o = (\mathbf{M}, \mathbf{F}) \in \Delta$}
			\State Sample pairs $s = (succ, \, t_{\text{fail}})$
			\State $p'_{\text{blame}} \leftarrow$ \Call{Bayes}{$p_{\text{blame}}$, $o$, $a$, $s$, $t_{\text{fail}}$}
			\State Compute $H \left[ p'_{\text{blame}} \right]$ and add to $H_{\text{sam}}$
		\EndFor
		\State Compute $\mathbf{E} \left[ H \left( p_{\text{blame}} \left( f \, | \, \dots o, \, a \right) \right) \right]$ from $H_{\text{sam}}$
		\State $\mathbf{E}[I(a)] \leftarrow H \left[ p_{\text{blame}} \right] - \mathbf{E} \left[ H \left( p_{\text{blame}} \left( f \, | \, \dots o, \, a \right) \right) \right]$
	\EndFor
	\State \Return{$(a, \, I(a))$ with maximum $I(a)$}
\EndFunction
\Function{Bayes}{$p_{\text{blame}}\left( f \, | \, o_{1:t}, \, a_{1:t} \right)$, $o$, $a$, $succ$, $t_{\text{fail}}$}
	\State Estimate $p_a \left( o \, | \, f, \, \Delta_a \right)$ according to equation (\ref{equ:actuallikelihood})
	\State \Return{$p_{\text{blame}}\left( f \, | \, o_{1:t}, \, o, \, a_{1:t}, \, a \right) \propto$\\\hfill $p_a \left( o \, | \, f, \, \Delta_a \right) \, p_{\text{blame}}\left( f \, | \, o_{1:t}, \, a_{1:t} \right)$}
\EndFunction
\end{algorithmic}
\end{algorithm}
\subsection{Training the Observation Model}
For the MOM we follow the idea of learning an
encoder/decoder neural network and using the reconstruction error to
determine whether a given sequence is part of the training
distribution or not. Similarly, the reconstruction error was used
for anomaly detection \cite{sakurada2014anomaly,marchi2015novel}.

\begin{figure}
  \centering
  \includegraphics[width=0.3\textwidth]{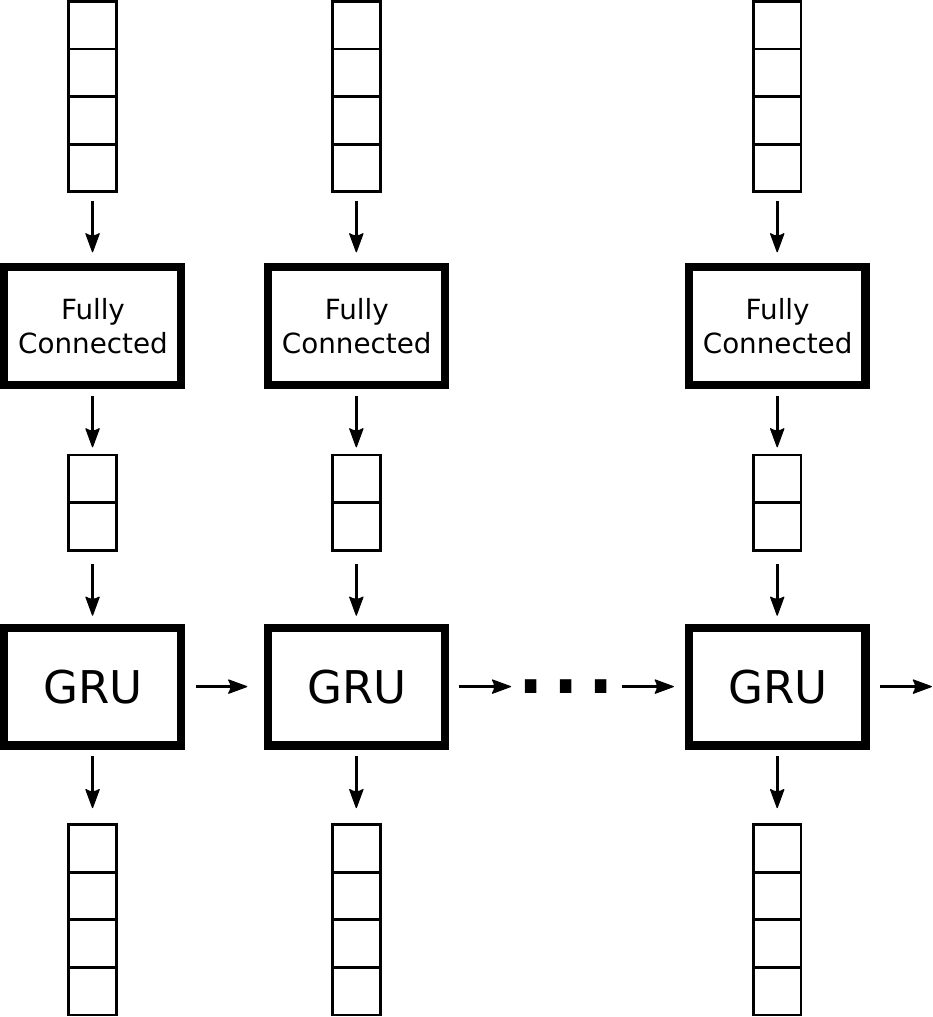}
  \caption{Schematic of the network architecture used for the
    measurement observation model\label{fig:netarch}}
\end{figure}

Our neural network is implemented as follows: Each vector of the time
series is encoded by the same fully connected neural network with
fewer output neurons than there are dimensions in the input vector.
This inevitably means that information is lost. This compression step
is followed by a layer of Gated Recurrent Units (GRUs)
\cite{cho2014properties}, with a number of output neurons equal
to the dimensions of the input vector. Fig.~\ref{fig:netarch} shows
a schematic sketch of the network used.
The network is trained by providing a given time series as input, and
the same time series as expected output.
As a loss function we use the average of the cosine similarities
between the vectors of the actual
output of the network and the expected output (the original time series).
This way, the network is forced to compress the time
series and reconstruct a time series as close to the original as
possible.
The cosine similarity is later also used as a metric for the reconstruction error.
The network is trained end to end using ADAM as an optimization scheme
\cite{kingma2014adam}.
Since GRUs have recurrent connections, they can use information from
previously-seen vectors in the reconstruction of following vectors.
Therefore, the decoding is done as a whole for the complete time
series, although the encoding is done for each vector separately.

After such a network is trained on multiple time series of successful
skill executions, we can assume that it has specialised to encode and
decode sequences of this kind with a low reconstruction error. The
hypothesis is that sequences deviating from successful examples will
have a higher reconstruction error.
Unfortunately, depending on the different phases of execution, the
reconstruction error will also vary for the successful examples. In
practice this means that there are times at which a high
reconstruction error is more suspicious. To incorporate this
information in our model, we calculate the mean and standard deviation
of the reconstruction errors of all successful examples for each time
step. This gives us a normal distribution per time step, which we
use to determine the likelihood that a given reconstruction error is
of a familiar time series. For a given reconstruction error, we use
the probability density at this point as an indication of the
likelihood that this error is produced by a successful sequence. Since
we get a reconstruction error at each time step, we are able to detect
at which point in time the sequence begins to deviate from successful
examples. Therefore, we can infer at which point an error has
occurred. To filter out inevitable noise, we smooth the resulting
likelihood sequence using moving average.

Figure \ref{fig:momeval} shows the reconstruction error likelihood of
different sequences. It can be seen that the likelihood of
failing sequences (red) generally falls to a value close to 0
over time. The drop-off point is where an error in execution of the
task has likely occurred. The successful sequences used during training
are shown in green. Additional successful examples, not seen during
training, are shown in blue.
\subsection{Training the Functional Profiling Fingerprint}
We assume that fingerprints, i.e. the time-series of function calls,
are Gaussian distributed separately at each time step for a specific skill.
For each cell $fc_i(t)$ of the matrix $\mathbf{F}$ in equation~(\ref{equ:fingerprintmatrix}),
we estimate the mean and
variance over all sample executions of a skill $a \in A$ independently.
This is equivalent to approximating the fingerprint distribution by a
multivariate Gaussian distribution
\begin{equation}
	p_a \left( \mathbf{F} \, | \, \mathbf{M}, \, t, \, succ = \text{true} \right) \approx
\mathcal{N}\left( \mu_{\mathbf{F}}(t), \, \Sigma(t) \right)
\end{equation}
with a diagonal co-variance matrix $\Sigma(t)$.
Therefore, the mean and variance are computed for each time step independently, but
are averaged over all observed executions.
This allows us to model the fingerprint of each function $f$ separately,
i.e.\ the probability
\begin{dmath}
	p_a^f \left( fc \, | \, \mathbf{M}, \, t, \, succ = \text{true} \right) = \\
		\int_{t' \neq t}\int_{f' \neq f}{p_a \left( \mathbf{F} \, | \, \mathbf{M}, \, t, \, succ = \text{true} \right) \, d \, f' \, d \, t'}
\end{dmath}
of function $f$ being called at time $t$.
While the independence assumption is not true in general,
we assume a certain degree of smoothness for a specific
skill where the function call time-series is similar for all
supported environment states (e.g. also for closed-loop controllers).
This allows us to further approximate by averaging over all observaions $\mathbf{M}$.
We stress that, in contrast, the MOM demands a more powerful modelling technique such as
deep learning in order to model sensor data can vary strongly.
\subsection{Bayesian Inference for Bug Detection}
\label{sec:bayesian}
When a skill $a \in A$ is executed, an observation
$o = \left( \mathbf{M}_a^{\text{exec}}, \, \mathbf{F}_a^{\text{exec}} \right)$
is measured.
We seek to update the probability distribution
$p_{\text{blame}} \left( f = f_i \, | \, o_{1:T}, o, \, a_{1:T}, a \right)$ accordingly.
Using Bayes' theorem and assuming
that the likelihood function does not depend on the previous
actions $a_{1:T}$ and observations $o_{1:T}$, i.e.\
$p_a \left( o \, | \, f = f_i, \, o_{1:T}, \, a_{1:T}, \, \Delta_a \right) \approx
p_a \left( o \, | \, f = f_i, \, \Delta_a \right)$, we can write
\begin{dmath}
	p_{\text{blame}} \left( f = f_i \, | \, o_{1:T}, o, \, a_{1:T}, a \right) \propto \\
		p_a \left( o \, | \, f = f_i, \, o_{1:T}, \, a_{1:T} \, \Delta_a \right) \,
		p_{\text{blame}} \left( f = f_i \, | \, o_{1:T}, \, a_{1:T} \right) 
		\approx \\
		p_a \left( o \, | \, f = f_i, \, \Delta_a \right) \,
		p_{\text{blame}} \left( f = f_i \, | \, o_{1:T}, \, a_{1:T} \right).
\end{dmath}
The likelihood function $p_a \left( o \, | \, f = f_i, \, \Delta_a \right)$
denotes the probability of seeing a certain observation $o$ given that the function
$f = f_i$ is buggy.
For the sake of readability we omit the condition on the database $\Delta_a$
for the distribution $p_{\text{blame}}$.
The likelihood function typically is unknown but can be approximated
given the following assumptions:

(i) A failure in function $f_i$ executed at time $t_{\text{exec}}$
	influences
	$p_a \left( succ \, | \, \mathbf{M}, \, t = t_{\text{exec}} + \delta t \right)$
	with a probability proportional to
	$\mathrm{e}^{-\alpha \delta t} = \mathrm{e}^{-\alpha \left(t_{\text{fail}} - t_{\text{exec}}\right()}$,
	where $t_{\text{fail}}$ is the estimated failure time and $\alpha$ is a free parameter.
	
(ii) As all probabilities
	$p_a^{f_i} \left( fc \, | \, \mathbf{M}, \, t, \, succ = \text{true} \right)$
	for all $f_i$ are assumed to be independent, the failure of $f_i$ only
	causes changes in the $i$th row of $\mathbf{F}_a^{\text{exec}}$.

(iii) A failure of $f_i$ can, but does not have to, cause a change in the
	$i$th row of the fingerprint $\mathbf{F}_a^{\text{exec}}$, e.g. the execution length of $f_i$
	changes (which would affect $\mathbf{F}_a^{\text{exec}}$) or just the behaviour of $f_i$ changes
	(which does not mean that the distribution of function calls is affected).\\
	
A skill is executed and the MOM is used to estimate the earliest time step $t_{\text{fail}}$
with $p_a \left( succ \, | \, \mathbf{M}, \, t_{\text{fail}} \right) \leq p_{\text{thresh}}$.
We use the fingerprints $\mathbf{F}_a^{\Delta}$ in the database $\Delta$ of skill $a$
to compute the expected values of exponentially weighted function counts
$\mathrm{e}^{- \alpha \delta t} fc_i^{\Delta}(t)$ for each function $f_i$ with
\begin{dmath}
\mathbf{E}_{t_{\text{fail}}}^{\Delta} \left[ f_i \right] \hiderel{\coloneqq} \mathbf{E}_{t_{\text{fail}}}^{\Delta} \left[ \mathrm{e}^{- \alpha \delta t} fc_i \right] \hiderel{=} \\
\int_{\mathbb{N}} \int_{\mathbf{M}} \int_{0}^{t_{\text{fail}}}{\mathrm{e}^{- \alpha \delta t} \, fc \, p_a^{f_i} \left( fc \, | \, \mathbf{M}, \, t, \, succ = \text{true} \right)} \, \mathrm{d}t \, \mathrm{d}\mathbf{M} \, \mathrm{d}fc
\label{equ:expectedvalues}
\end{dmath}
and corresponding variances
$var_{t_{\text{fail}}}^{\Delta} \left[ f_i \right] \coloneqq
var_{t_{\text{fail}}}^{\Delta} \left[ \mathrm{e}^{- \alpha \delta t} fc_i \right]$.
The open parameter $\alpha$ determines the size of the time window on the fingerprint.
We further compute the weighted mean of the executed fingerprint $\mathbf{F}_a^{\text{exec}}$ with
\begin{equation}
	\mu_{t_{\text{fail}}}^{\text{exec}}[f_i] = \frac{1}{T} \sum_{t = 0}^{t_{\text{fail}}}{
	  \mathrm{e}^{- \alpha \delta t} \, fc^{\text{exec}}_i(t).
	}
\end{equation}
These values are used to compare the observed fingerprint of the executed skill $a$
to the corresponding FPF in the database $\Delta_a$.
The assumptions defined above culminate in
\begin{dmath}
	p_a \left( o \, | \, f = f_i, \, \Delta_a \right) =
	\begin{cases}
		p_{\text{dev}}			&	
									\text{if } a \text{ was successful}\\
		\frac{\left( 1 + p_{\text{dev}} \right)}{2}		
								&		\text{else,}
	\end{cases}
\label{equ:actuallikelihood}
\end{dmath}
\begin{equation}
	p_{\text{dev}} = \mathcal{N} \left(x \in \left[ \mu_{t_{\text{fail}}}^{\text{exec}}[f_i], \, \mathbf{E}_{t_{\text{fail}}}^{\Delta} \left[ f_i \right] \right] \,
		| \, \mathbf{E}_{t_{\text{fail}}}^{\Delta} \left[ f_i \right], \,
		var_{t_{\text{fail}}}^{\Delta} \left[ f_i \right] \right).
\label{equ:pdev}
\end{equation}
\begin{figure*}[t!]
  \centering
  \subfloat[Expected information gains for skills with simulated fingerprints
  $\mathbf{F}_{a_1} = (1, \, 2)$ (blue),
  $\mathbf{F}_{a_2} = (2, \, 4, \, 5)$ (red),
  $\mathbf{F}_{a_3} = (3, \, 4, \, 6)$ (green, behind magenta),
  $\mathbf{F}_{a_4} = (3, \, 4, \, 5, \, 6)$ (magenta).
  The skills $a_3$ and $a_4$ do not share any function with $a_1$, among which the error must be after executing $a_1$,
  and are ignored.
  For different executions of the algorithm the height and exact locations of the maxima might
  shift slightly due to the random generation of the fingerprint data.
  ]
  {\label{fig:sim1igs}\includegraphics[width=0.44\textwidth]{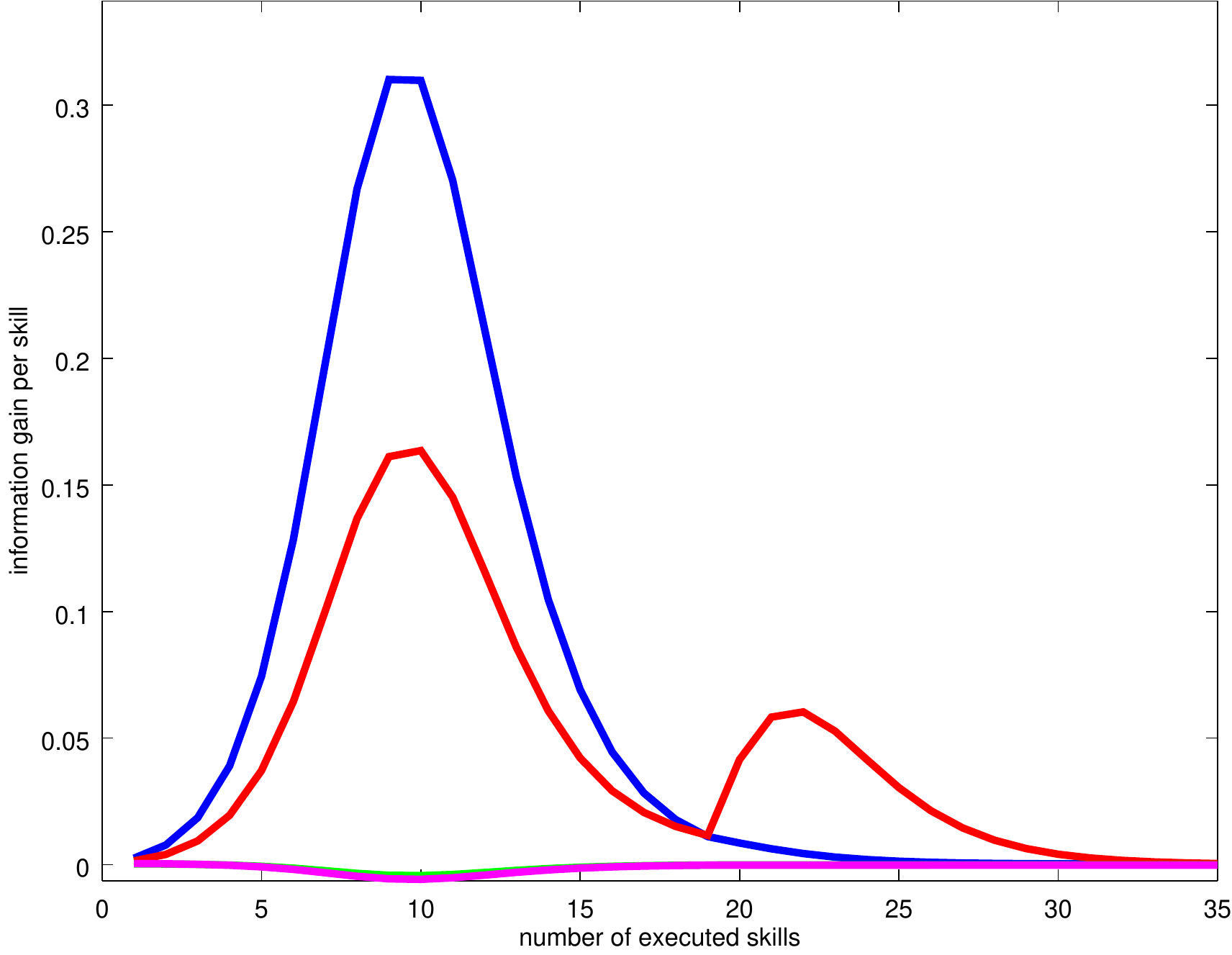}}
  \qquad
  \subfloat[$p_{\text{blame}}$ for the first 6 functions.
All other functions have a low probability (comparable to the magenta curve)
and are omitted. Initially the dominant skill is $a_1$ (cf.\ Fig.~\ref{fig:sim1igs}), which cannot discriminate between functions $f_1$ (red) and $f_2$ (blue).
After 19 steps, skill $a_2$ is chosen which identifies $f_2$ as the cause of error.
]{\label{fig:sim1pblame}\includegraphics[width=0.44\textwidth]{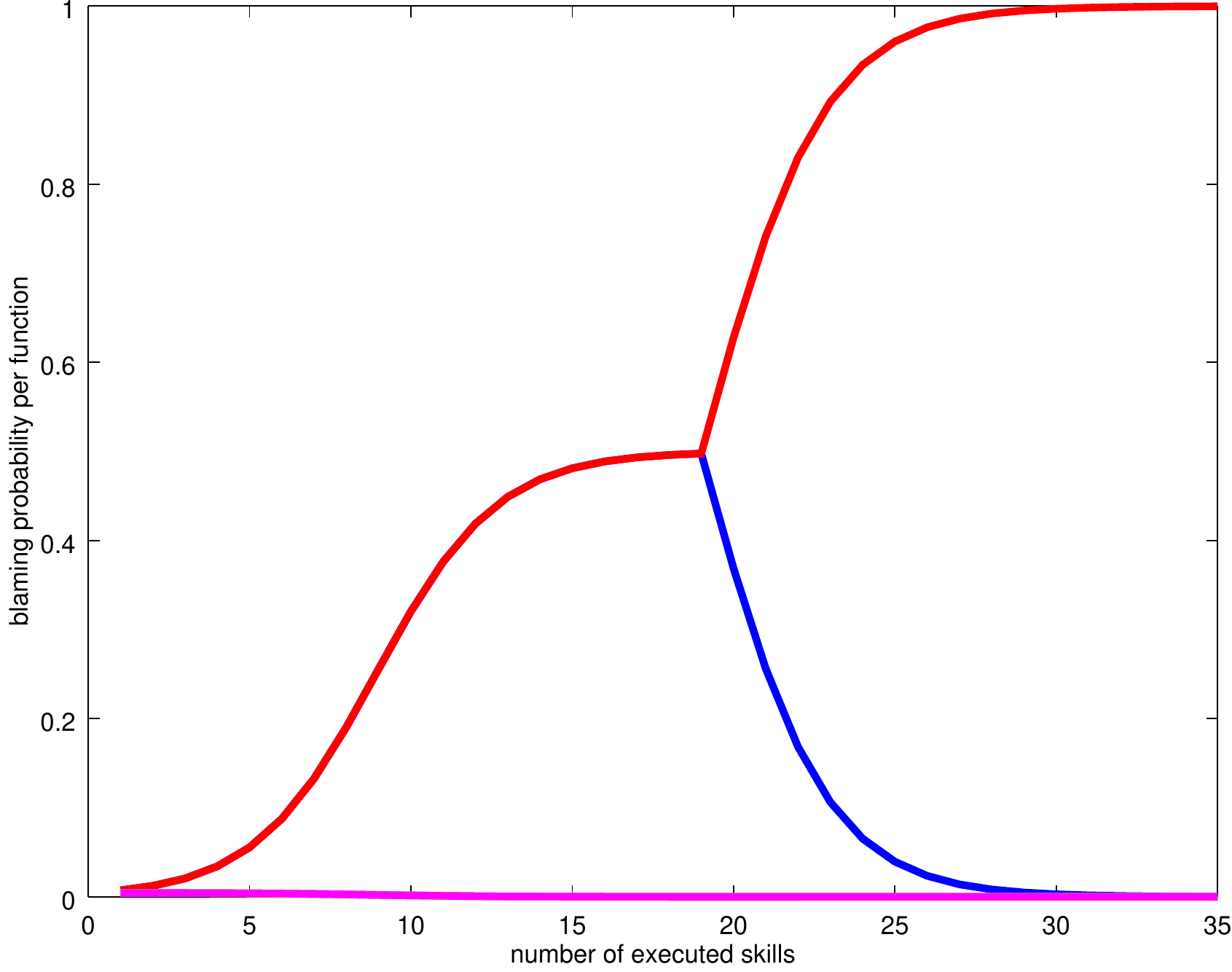}}
\caption{Typical information gains for four different skills and the respective probabilities $p_{\text{blame}}$ for functions 1 to 6 plotted
over the number of executed skills.
The function $f_2$ is simulated to be the cause of the error.
}
\label{fig:simcase3}
\end{figure*}
\begin{figure}
\center
\includegraphics[width=0.44\textwidth]{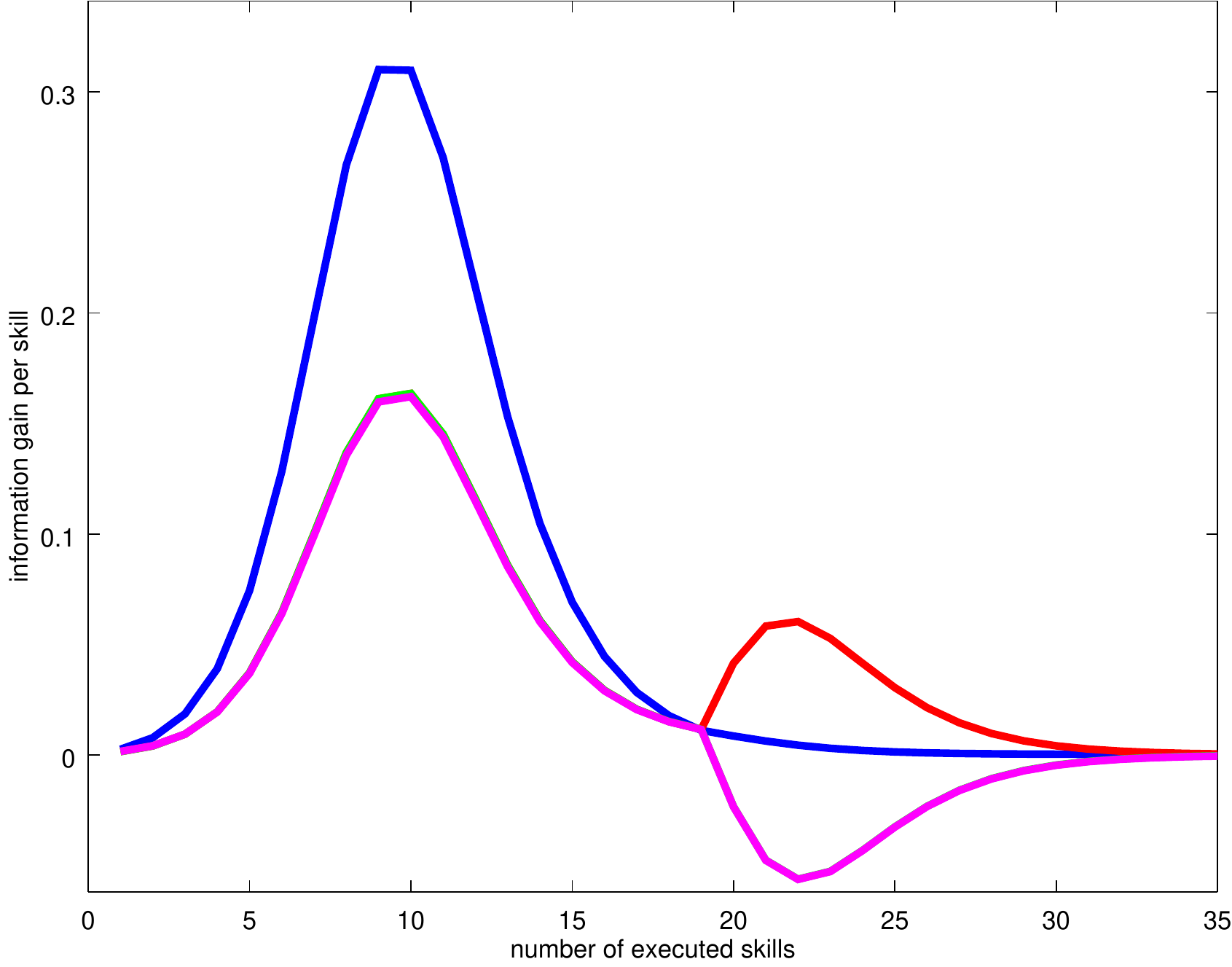}
\caption{Expected information gains for skills with simulated fingerprints
  $\mathbf{F}_{a_1} = (1, \, 2)$ (blue),
  $\mathbf{F}_{a_2} = (2, \, 4, \, 5)$ (red, hidden behind magenta),
  $\mathbf{F}_{a_3} = (1, \, 3, \, 6)$ (green, hidden behind magenta),
  $\mathbf{F}_{a_4} = (1, \, 3, \, 4, \, 6)$ (magenta).
As opposed to the scenario in Fig.~\ref{fig:sim1igs}, the skills $a_3$ and $a_4$
can also help to identify the bug.
The skills $a_2$ and $a_3$ are equally likely to be helpful and have the same
expected information gain, whereas skill $a_4$ uses one function more than $a_3$
and therefore has a slightly lower expected information gain.
}
\label{fig:sim3igs}
\end{figure}
The probability $p_{dev}$ measures how much a measurement
$\mu_{t_{\text{fail}}}^{\text{exec}}$ deviates from a regular
execution of the skill $a$ within the time window
$t \in \left[ 0, \, t_{\text{fail}} \right]$.
Case 1 of equation \ref{equ:actuallikelihood} treats the case
of a successful execution.
The probability of observing a successful observation $o$
given the function $f_i$ has a bug should be low.
However, a function might still have a bug but just did not
affect the skill enough to make it fail.
Therefore, $p_a \left( o \, | \, f = f_i, \, \Delta_a \right)$
should increase if the finger print strongly differs
from other successful experiences.
The second case of equation \ref{equ:actuallikelihood} applies
if the skill was not executed successfully.
Each executed function should at least be suspicious, i.e.
$p_a \left( o \, | \, f = f_i, \, \Delta_a \right) \geq 1 / 2$.
The probability increases proportionally to the deviation from
previous successful experiences.
\subsection{Skill Selection by Information Gain Maximisation}
\label{sec:informationgain}
The previous section described how to update the belief
$p_{\text{blame}} \left( f = f_i \, | \, o_{1:T}, o, \, a_{1:T}, a \right)$
given a skill $a$ was selected.
This skill should be selected such that the entropy of the probability
distribution $p_{\text{blame}}$ decreases.
A common way to solve this problem is to maximise the expected information gain.
The information gain is defined by
\begin{dmath}
	I \left[ a \right] \hiderel{=}
	H \left[ p_{\text{blame}} \left( f \hiderel{=} f_i \, | \, o_{1:T}, \, a_{1:T} \right) \right]
	- 
	H \left[ p_{\text{blame}} \left( f \hiderel{=} f_i \, | \, o_{1:T}, o, \, a_{1:T}, a \right) \right]
\end{dmath}
Before executing the skill $a \in A$, the information gain cannot be computed directly,
as the corresponding observation $o$ is not known yet.
However, the current belief
$p_{\text{blame}} \left( f = f_i \, | \, o_{1:T}, \, a_{1:T} \right)$
can be used to estimate the expected information gain
$\mathbf{E} \left[ I \left[ a \right] \right]$.
We estimate $\mathbf{E} \left[ I \left[ a \right] \right]$ by uniformly sampling
pairs $\left( succ, \, t_{\text{fail}} \right)$ for each sample
in the database $\Delta_a$ with random success
$succ \in \{ \text{true}, \, \text{false} \}$ and random failure time
with probability $p(t_{\text{fail}}) = 1 / T$.
At each step a Bayesian belief update is performed, and
the expected entropy
$\mathbf{E} \left[ H \left[ p_{\text{blame}} \left( f \hiderel{=} f_i \, | \, o_{1:T}, o, \, a_{1:T}, a \right) \right] \right]$
is estimated.
The current entropy
$H \left[ p_{\text{blame}} \left( f \hiderel{=} f_i \, | \, o_{1:T}, \, a_{1:T} \right) \right]$
can be computed in closed form and the robot optimises
\begin{dmath}
	a_{\text{next}} \hiderel{=} \argmax_{a}{\mathbf{E} \left[ I \left( a \right) \right]} \hiderel{=} 
	\argmax_{a}{}\\H \left[ p_{\text{blame}} \left( f \hiderel{=} f_i \, | \, o_{1:T}, \, a_{1:T} \right) \right]
	- \mathbf{E} \left[ H \left[ p_{\text{blame}} \left( f \hiderel{=} f_i \, | \, o_{1:T}, o, \, a_{1:T}, a \right) \right] \right].
\end{dmath}
\section{Experiments}
We evaluate our method in simulation and in real-world tasks.
In simulation we analyse the behaviour of the Bayesian inference
and the information gain optimisation independently of the MOM.
Our approach is tested with a real robot by implementing a set of
purposefully sabotaged skills.
\subsection{Simulated Experiments}
\begin{figure*}[t!]
  \centering
  \subfloat[Expected information gains for skills with simulated fingerprints $\mathbf{F}_{a_1} = (1, \, 2)$ (blue),
  $\mathbf{F}_{a_2} = (2, \, 4)$ (red, not visible, aligned with blue),
  $\mathbf{F}_{a_3} = (1, \, 3, \, 6)$ (green, not visible, aligned with magenta),
  $\mathbf{F}_{a_4} = (1, \, 3, \, 4, \, 6)$ (magenta).
  Skills $a_1$ and $a_2$ have about the same expected information gain, which causes alternate execution.
  ]{\label{fig:sim2igs}\includegraphics[width=0.44\textwidth]{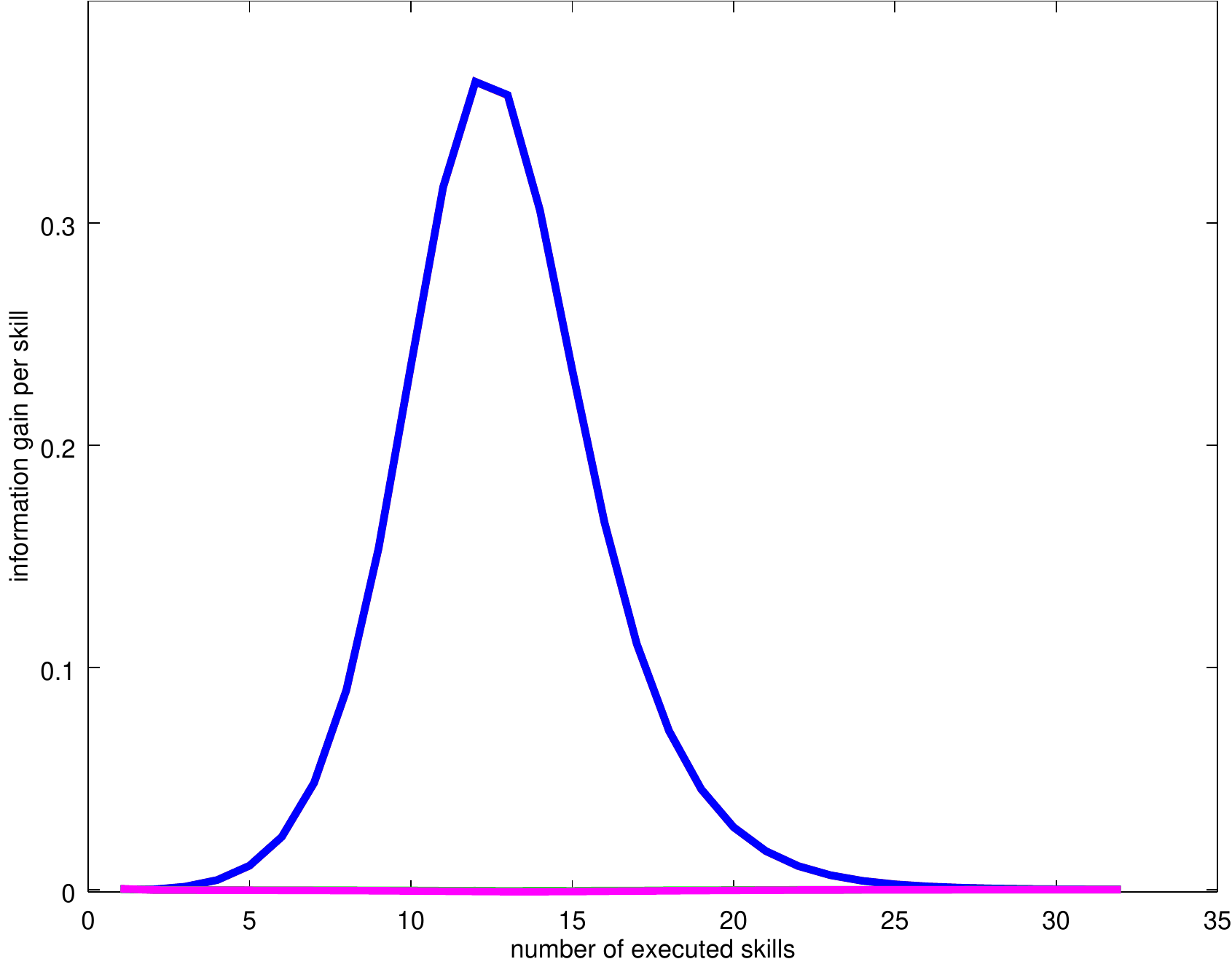}}
  \qquad
  \subfloat[The failure function is correctly identified early.
  The alternate execution of skills $a_1$ and $a_2$ yields a continuous improvement of confidence.
  All probablities excecpt for $f_2$ (blue) are close to 0.]{\label{fig:sim2pblame}\includegraphics[width=0.44\textwidth]{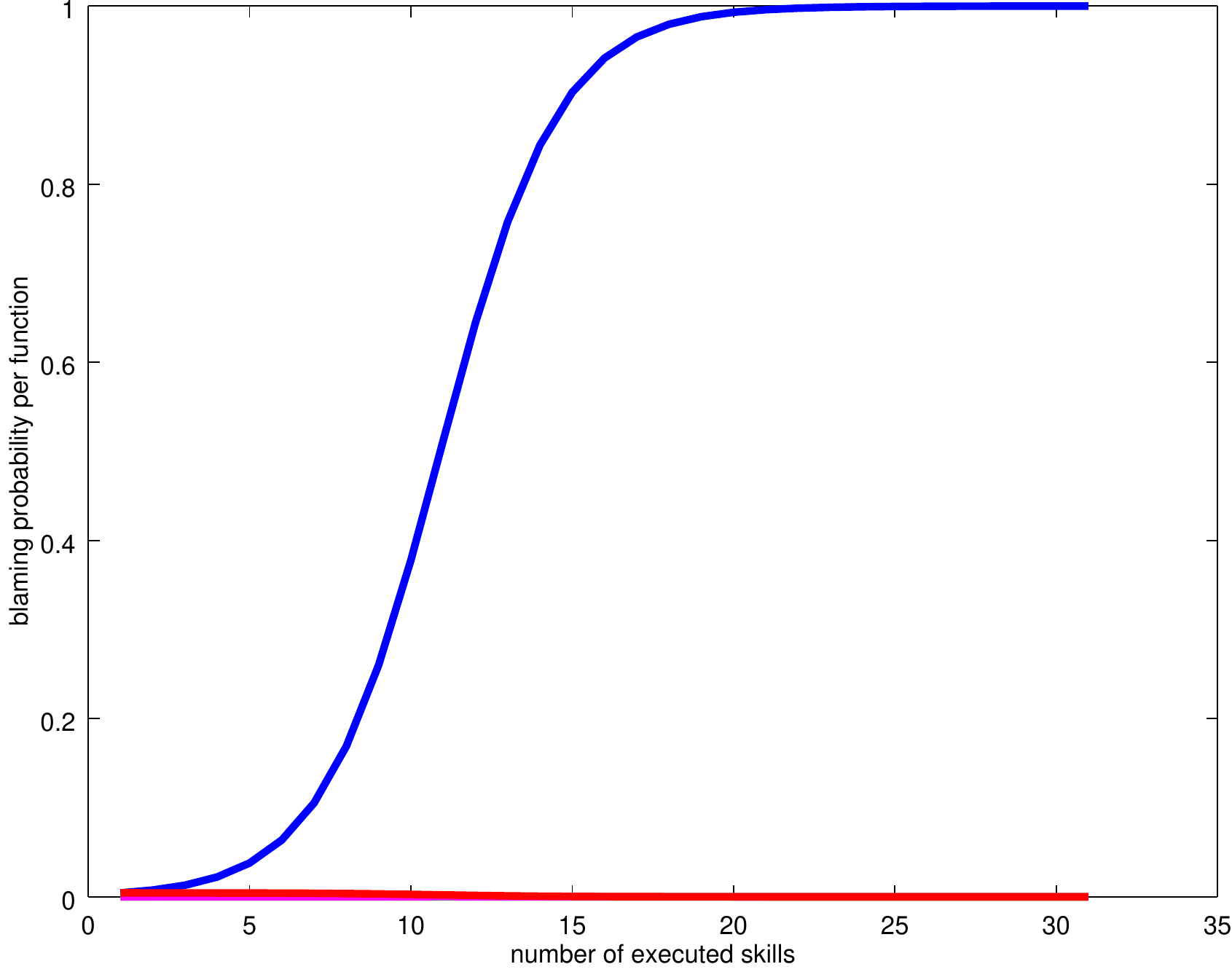}}
\caption{Degenerate scenario: information gains for four different skills and the respective probabilities $p_{\text{blame}}$ for functions 1 to 6.
The skills $a_1$ and $a_2$ are executed alternately.
The function $f_2$ is simulated to be the cause of the error.}
\label{fig:simcase2}
\end{figure*}

The Bayesian reasoning is decoupled from the MOM by generating artificial
fingerprints of the type $\mathbf{F}_a \left( f_1, \, f_2, \, \dots \, f_F \right) = \left( \mathbf{fc}_1, \, \mathbf{fc}_2, \, \dots, , \, \mathbf{fc}_F \right)^T$.
If $f_i = \text{false}$, the vector $\mathbf{fc}_i$ is chosen with $\mathbf{fc}_i = \vec{0}$.
Otherwise it is given by $\mathbf{fc}_i = \left( fc_i(0) \propto \mathcal{N}(\mu_i, \sigma_i), \, \dots , fc_i(T) \propto \mathcal{N}(\mu_i, \sigma_i) \right)^T$.
If the matrices $\mathbf{F}_a \left( f_1, \, f_2, \, \dots \, f_F \right)$ are generated this way,
the particular choice of the MOM is irrelevant.
Artificial skills using the first 6 out of 241 available functions were generated.
In order to simplify the notation, only used functions are denoted, e.g.
$\mathbf{F}_a \left( \text{true}, \, \text{true}, \, \text{false}, \, \dots, \, \text{false} \right)
\equiv \mathbf{F}_a \left( 1, \, 2 \right)$.
In Figs.\ \ref{fig:simcase3}~--~\ref{fig:simcase2} typical evolutions of information gains
and blaming probabilities by using skills with different fingerprints are shown.
In all scenarios a bug in function $f_2$ causes an
error if a skill uses $f_2$.
Our system was able to detect the error in all scenarios with $p_{\text{blame}}(f = f_2) \approx 1$.
Fig. \ref{fig:simcase3} shows a typical scenario, in which one skill is executed until
another skill has a higher information gain.
The skill is switched after 19 executions in order to discriminate
between two failure candidates $f_1$ and $f_2$.
Note that all functions except $f_1$ and $f_2$ have almost 0 probability
as after executing the skill $a_1$ has identified that a bug must be in functions $f_1$ or $f_2$
which places the algorithm in a local optimum.
If further bugs are contained in the software, the system has to be re-iterated when the
bug in $f_2$ is fixed.
The high number of functions with very low probability causes the probabilities
in Fig.~\ref{fig:sim1pblame} to look like they are not normalised, however,
this is just a visual artifact.
Fig.~\ref{fig:sim3igs} shows a similar scenario with a slightly different selection of
used functions per skill.
Even though the information gains develop differently, the bug belief develops the same
way as in Fig.~\ref{fig:sim1pblame} because no different actions are taken.
Fig.~\ref{fig:simcase2} shows a degenerate case in which two skills contribute equally much
to identify the function $f_2$ and are executed alternately while the confidence in
$f_2$ increases continuously.
Even though the correct function is identified, the skill switching might require extra
effort for preparation of different environments.
\begin{figure*}[t!]
  \centering
  \subfloat[Simple grasping skill]{\label{fig:momgrasp}\includegraphics[width=0.48\textwidth]{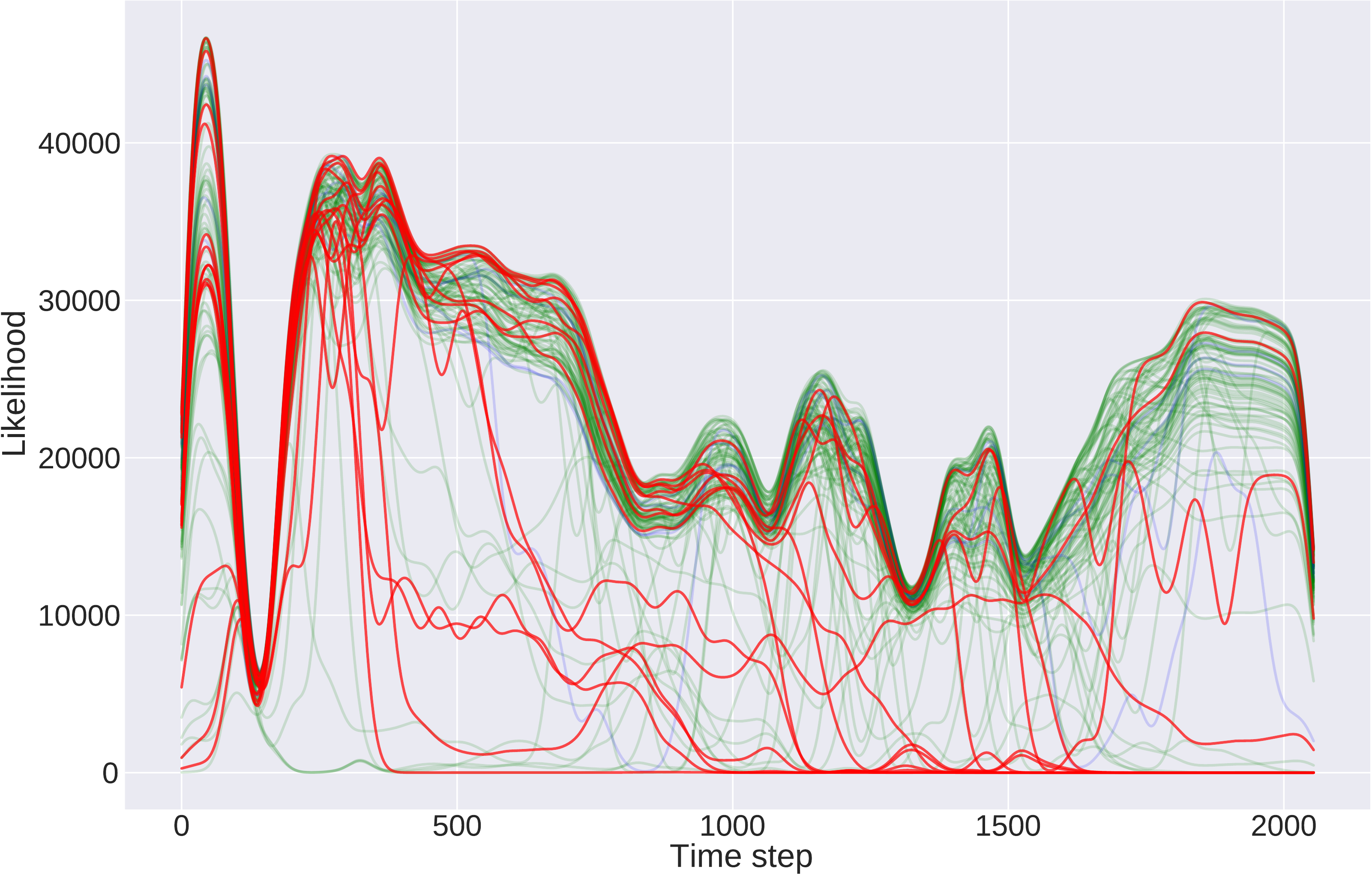}}
  \qquad
  \subfloat[Handover skill]{\label{fig:momhandover}\includegraphics[width=0.48\textwidth]{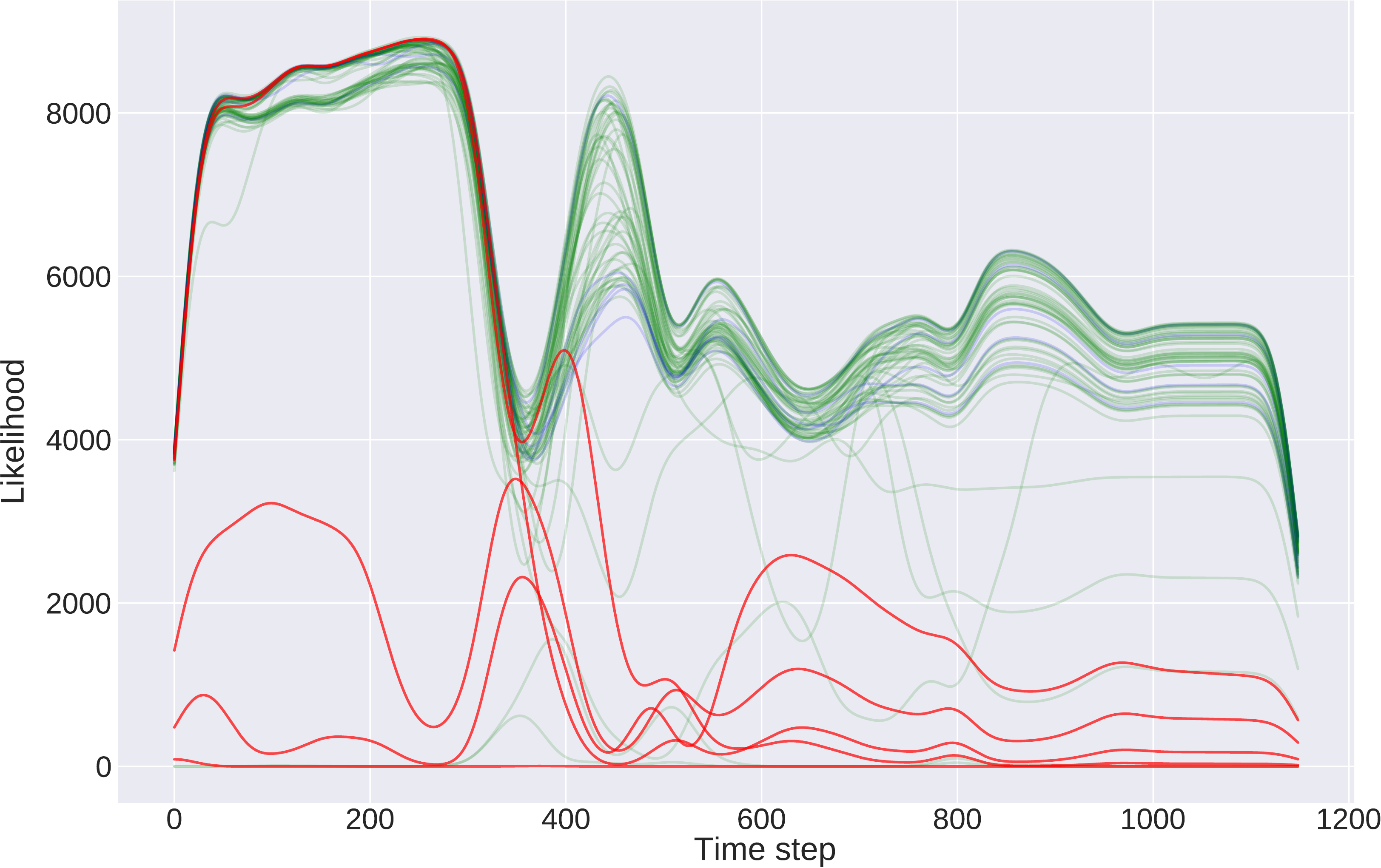}}
  \caption{MOM in two simple skill scenarios. The graphs show the
    reconstruction error likelihood over the execution time. It
    includes the (only positive) training samples (green), the
    positive test samples (blue) and the negative test samples (red).
    As can be seen, the method does not work in all cases, but the
    likelihood of a correct execution of a skill being classified as
    not working is relatively small.}
\label{fig:momeval}
\end{figure*}
\subsection{Real-world experiments}
\begin{figure}
  \centering
  \includegraphics[width=0.49\textwidth]{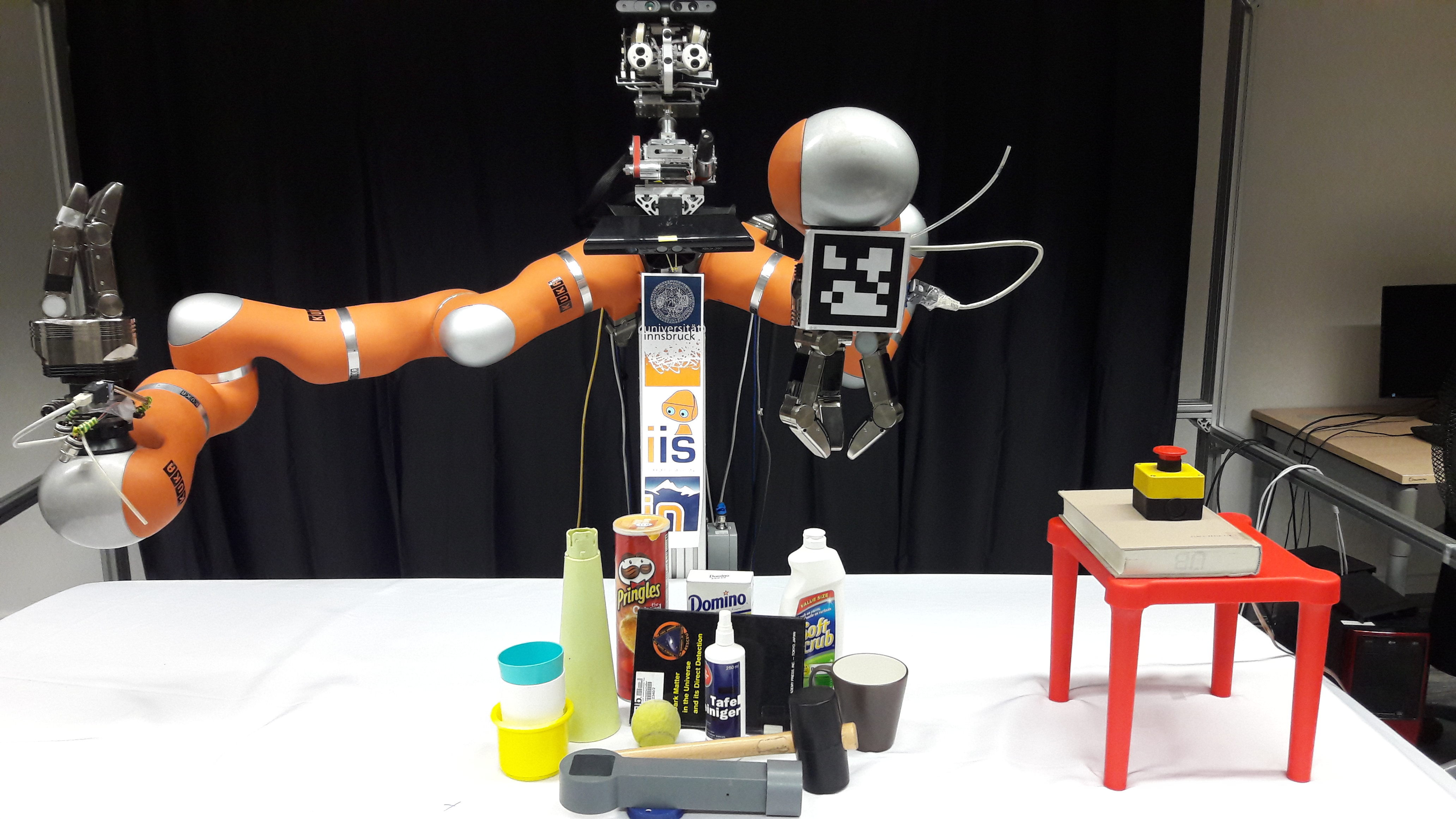}
  \caption{Robot setting and the used objects.}
    \label{fig:setting}
\end{figure}

We show the applicability of deep learning to failure detection in
two simple tasks: simple grasping and object handover.
Further, our system is tested with a set of simple skills
by purposefully introducing bugs to the software.
The robot setting used can be found in Fig.~\ref{fig:setting}.
It consists of two KUKA LWR 4+ robotic arms with one Schunk SDH gripper
attached to each arm.
The KUKA arms provide a control interface
called \emph{fast research interface} (FRI), which enables control
and sensor data retrieval (joint positions, joint forces,
Cartesian forces and torques).
For object detection two Kinects are mounted - one on the chest and one above the robot.
Objects are localised by segmenting them from the table surface
and fitting a box to the remaining point cloud image by using PCL.
Three different skills are implemented:
\begin{itemize}
	\item Simple grasp: Objects are placed in front of the robot.
	A PCL based localiser recognises the objects and a Cartesian planner
	is used to move the end-effector.
	The fingers are closed and the object is lifted.
	\item Pressing a button: A red emergency button is placed in a fixed
	robot-relative position.
	A joint plan is executed and the button is pressed with the wrist and
	without using the fingers.
	\item Handover: The robot reaches forward by using a joint plan
	and waits until an object is placed in the hand.
	This event is detected by observing the end-effector forces.
	When an object is handed over, the fingers are closed.
\end{itemize}
\subsubsection{Evaluation of the MOM}
A set of erroneous executions was generated by either adding bugs to the code or by manually
interfering with the environment, e.g.\ by kicking the object out of the robots hand.

Our implementation of the MOM uses a fully connected neural network
with 32 output neurons and a RELU nonlinearity. The size of this
bottleneck depends on the dimensionality of the input vectors. In our
case, it was determined empirically. The GRU layer uses an internal
$tanh$-nonlinearity and a logistic sigmoid nonlinearity for the
output. The network is trained for 500 epochs which, on an nVidia GTX
1080 GPU for 70 sequences of 2000 time steps each, takes about five
minutes.

Fig.~\ref{fig:momgrasp} shows the MOM performance on the dataset for simple grasping.
For most negative samples the likelihood 
of being a successful sample drops close to zero, while this is not
the case for successful samples.
Two different failure types are visible: the first failure type arises very early, at $t < 500$, and corresponds to failed path planning which causes
the arm not to move at all.
When the arm moves, a failure cannot be detected as there is no contact with the object yet.
Failures are detected when contact is made with the object from time $t > 1000$.
In the handover task (Fig~\ref{fig:momhandover}), the failure is detected when the arm does
not move at all during the reaching motion or when the object should be placed in the hand, but is not. It should be noted that as soon as an error is detected, the further development of the reconstruction error is not relevant anymore.
\subsubsection{Running the complete system}
Sensor data and fingerprints were measured for at least 70 executions per skill.
In total, more than 80 functions (including robot control) were used by the skills.
Different types of bugs were purposefully introduced to the software.
To reduce the computational effort for the estimation of the information gain,
the decaying expected value in equation \ref{equ:expectedvalues}
was computed only for a window of 2 seconds before $t_{\text{fail}}$.

In the first scenario, the Cartesian planner was destroyed by introducing a constant shift.
This type of error could happen if the robot model is not correct.
The robot started to test the grasping skill which failed due to missing the object.
It identified a set of potentially faulty functions and the 4 functions with the highest
belief were \{\emph{getRobotId, planJointTrajectory, storeJointInfoToDatabase, updateFilter}\}.
None of these functions are related to the Cartesian shift bug.
However, the confidence of the robot was low and
\emph{planCartesianTrajectory} was included with a similar probability.
The grasping skill was used to confirm the belief until the button pressing skill provided
a higher expected information gain.
The button pressing skill was executed successfully because only a joint plan was used.
This allowed the robot to exclude most wrong guesses and to suggest the list
\{\emph{cartesianPtp, planCartesianTrajectory, computeIk, closeHand}\} with high probability.
The \emph{localiseObject} is not suspicious as it was executed outside the time
window, but is in the list as well if the window is enlarged.
The \emph{handover} skill enabled the robot to eliminate \emph{closeHand}.
All identified functions, i.e. \{\emph{cartesianPtp, planCartesianTrajectory, computeIk}\},
are involved in Cartesian planning (and not in joint planning).

In the second scenario commands for moving the hand were not forwarded to the hardware.
This affected the grasping skill and the handover skill, whereas the button pressing
skill succeeded.
All functions used in the button pressing skill were removed from the list of candidates
(e.g.\ for joint planning, arm control functions) and the functions for Cartesian
planning and hand control remained:
\{\emph{closeHand, cartesianPtp, planCartesianTrajectory, computeIk}\}.
In the next step the \emph{handover} skill was executed.
Because the Cartesian planning functions were not used in the failing
\emph{handover} skill, the corresponding functions were eliminated.

In another scenario, the localisation system was destroyed by returning a constant position.
In this case, the error is hard to detect:
In the grasping scenario, the robot misses the object long after the localiser was run.
If the time window is long enough, the system still identifies the localiser among
other plausible candidates and returns the set \{\emph{localiseObject, cartesianPtp,
planCartesianTrajectory, computeIk}\} with high probabilities.
Given the three provided skills, the robot has no possibility to discriminate between
errors in the respective functions.
In our setting, the localiser and Cartesian planning are always
used together and an error in the localiser does not appear earlier
in the sensor data.
A video demonstration of the approach switching from one skill to another can be found online%
\footnote{\url{https://iis.uibk.ac.at/public/shangl/iros2017/hangl-iros2017.mp4}}.
\section{Conclusion and Future Work}
We introduced a skill-centric software testing approach that uses data
collected over the life-time of a robot, which eliminates the need
for defining separate test cases.
We use two different types of data: sensor data of successful skill
executions and corresponding profiling data.
We train a so-called measurement observation model (MOM) (deep learning)
and a functional profiling fingerprint (FPF) (Multivariate Gaussian model).
Skills are executed as test cases and sensor data is compared to previous experiences.
We use Bayesian belief updates to estimate a probability distribution
of which functions contain bugs.
The skills are selected in order to maximize the expected information gain.
The approach is evaluated in simulation and in real robot experiments
by purposefully introducing bugs to existing software.

This work discusses the problem of bug detection and autonomous testing.
Future work will be concerned with developing autonomous strategies
for bug fixing (e.g. by automatically performing
selective git roll-backs or autonomously replacing hardware).
Further, due to the lack of data, the
likelihood function of the Bayesian belief update (equation \ref{equ:actuallikelihood})
had to be hand-crafted following reasonable assumptions.
In a more general framework the likelihood function can be
trained automatically by providing ground truth data whenever
the programmer has fixed a certain bug.

\addtolength{\textheight}{-7cm}   




\section*{ACKNOWLEDGMENT}
This work has received funding from the European Union’s
Horizon 2020 research and innovation program under grant agreement no. 731761, IMAGINE.

\bibliographystyle{IEEEtran}
\bibliography{bibfile}

\emph{•}
\end{document}